\documentclass{article}
\usepackage{spconf,amsmath,graphicx}
\usepackage{color,hyperref}
\usepackage{amsmath}

\usepackage{amssymb}

\usepackage{algorithm}
\usepackage{algorithmicx}
\usepackage{algpseudocode}
\usepackage{multirow}
\usepackage{booktabs} 
\usepackage{url}
\usepackage{bm}
\usepackage{verbatim}
\usepackage{breakurl}
\usepackage{bbding}
\usepackage{tabu}

\newlength\savedwidth

\title{Overcoming Language Bias in Remote Sensing Visual Question Answering via Adversarial Training}
\name{Zhenghang Yuan, Lichao Mou, Xiao Xiang Zhu}
\address{ Data Science in Earth Observation, Technical University of Munich (TUM), Munich, Germany}
\begin{document}
%\ninept
%
\maketitle

\begin{abstract}
%\textcolor{red}{ 
The Visual Question Answering (VQA) system offers a user-friendly interface and enables human-computer interaction. However, VQA models commonly face the challenge of language bias, resulting from the learned superficial correlation between questions and answers. To address this issue, in this study, we present a novel framework to reduce the language bias of the VQA for remote sensing data (RSVQA). Specifically, we add an adversarial branch to the original VQA framework. Based on the adversarial branch, we introduce two regularizers to constrain the training process against language bias. Furthermore, to evaluate the performance in terms of language bias, we propose a new metric that combines standard accuracy with the performance drop when incorporating question and random image information. Experimental results demonstrate the effectiveness of our method. We believe that our method can shed light on future work for reducing language bias on the RSVQA task.

\end{abstract}
\section{Introduction}
\label{sec:intro}
%vqa这个任务是一个很好的让用户理解自然图像并交互的一个任务。rsvqa为终端用户提供了可交互性
The volume of remote sensing data is increasing significantly on a daily basis, offering convenience and support for applications, such as urban planning, disaster assessment, and environmental monitoring \cite{xiong2022earthnets, 9184118, xiong2023gamus}. However, remote sensing images are not as easily comprehensible as natural images. Many ordinary users find it challenging to intuitively understand the details in remote sensing data. If natural language is used to depict the content of images, it will greatly increase the comprehensibility of remote sensing imagery \cite{yuan2023multilingual}. The visual question answering (VQA) task was proposed in \cite{antol2015vqa} and it employs natural language as an interface to bridge the gap between images and users. 

\begin{figure}[t!]
\includegraphics[width=0.5
\textwidth]{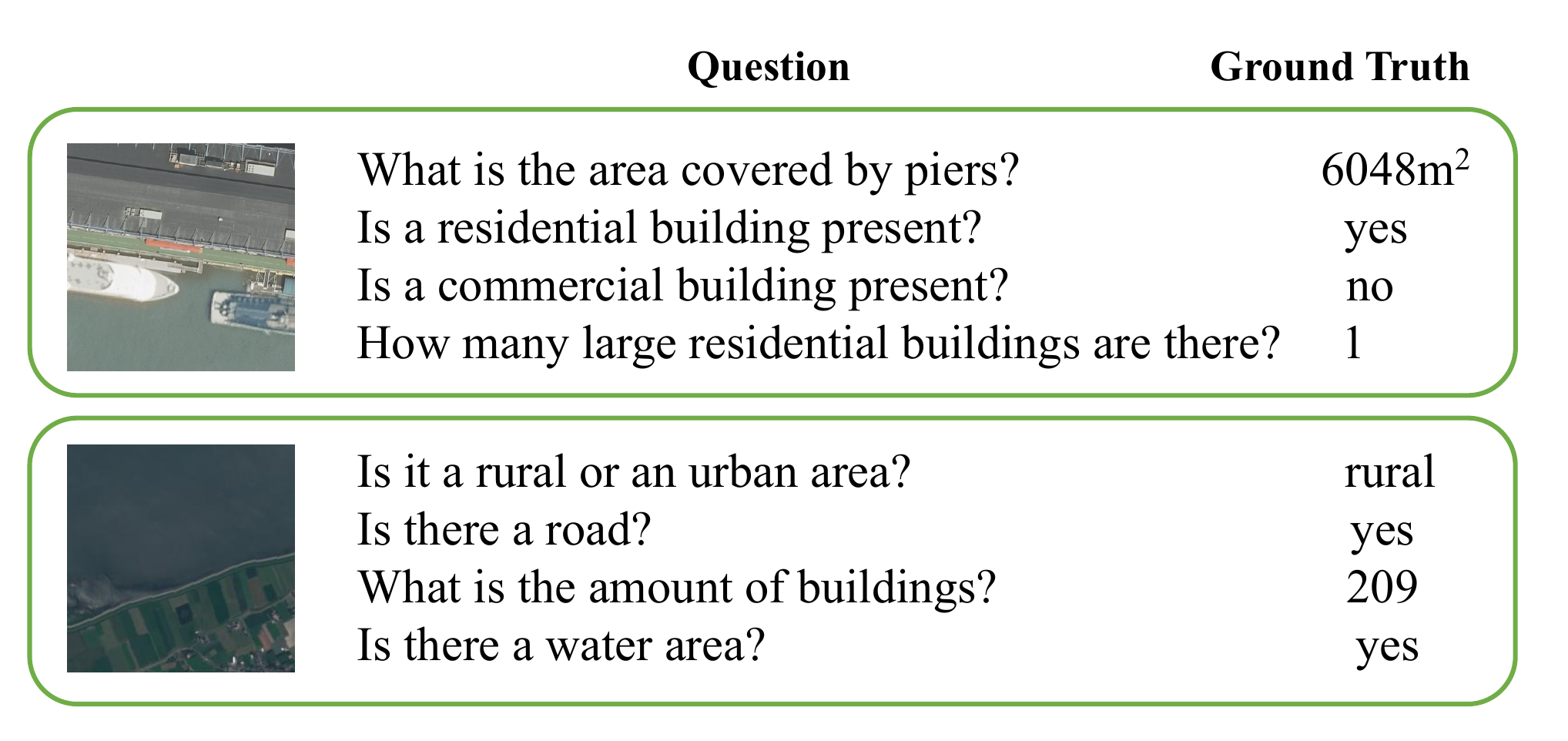}
\centering
\caption{RSVQA samples from high resolution (HR) and low resolution (LR) datasets.
}
\label{fig0}
\end{figure}

VQA for remote sensing data (RSVQA) is a recently trending research topic. As shown in Fig. \ref{fig0}, given an image and a question, the objective of RSVQA is to output the corresponding answer. This task was first introduced in the work \cite{lobry2020rsvqa}, where an RSVQA system was proposed to enable the extraction of information from remote sensing data. This work employed an automated approach to generating questions by utilizing information from OpenStreetMap, and constructed two datasets. Furthermore, convolutional neural networks (CNNs) and recurrent neural networks (RNNs) are used to learn visual and linguistic features from the given image and question, respectively. Subsequent to this study, many RSVQA works emerged.
Taking into account the varying levels of difficulty in questions for each image, Yuan et al. \cite{yuan2022easy} proposed a self-paced curriculum learning-based RSVQA model, where networks are trained with samples in an easy-to-hard manner. Chappuis et al. \cite{chappuis2022prompt} proposed the Prompt–RSVQA, which involves translating visual information into words and then utilizing them in a language-only model.
To offer users flexible access to change information, change detection-based VQA was introduced \cite{yuan2022change}. In this task, the input consists of an image pair and a change-related question, and the output is the corresponding answer. Additionally, this study proposed a change enhancing module for multitemporal feature encoding.

\begin{figure*}
\includegraphics[width=0.7\textwidth]{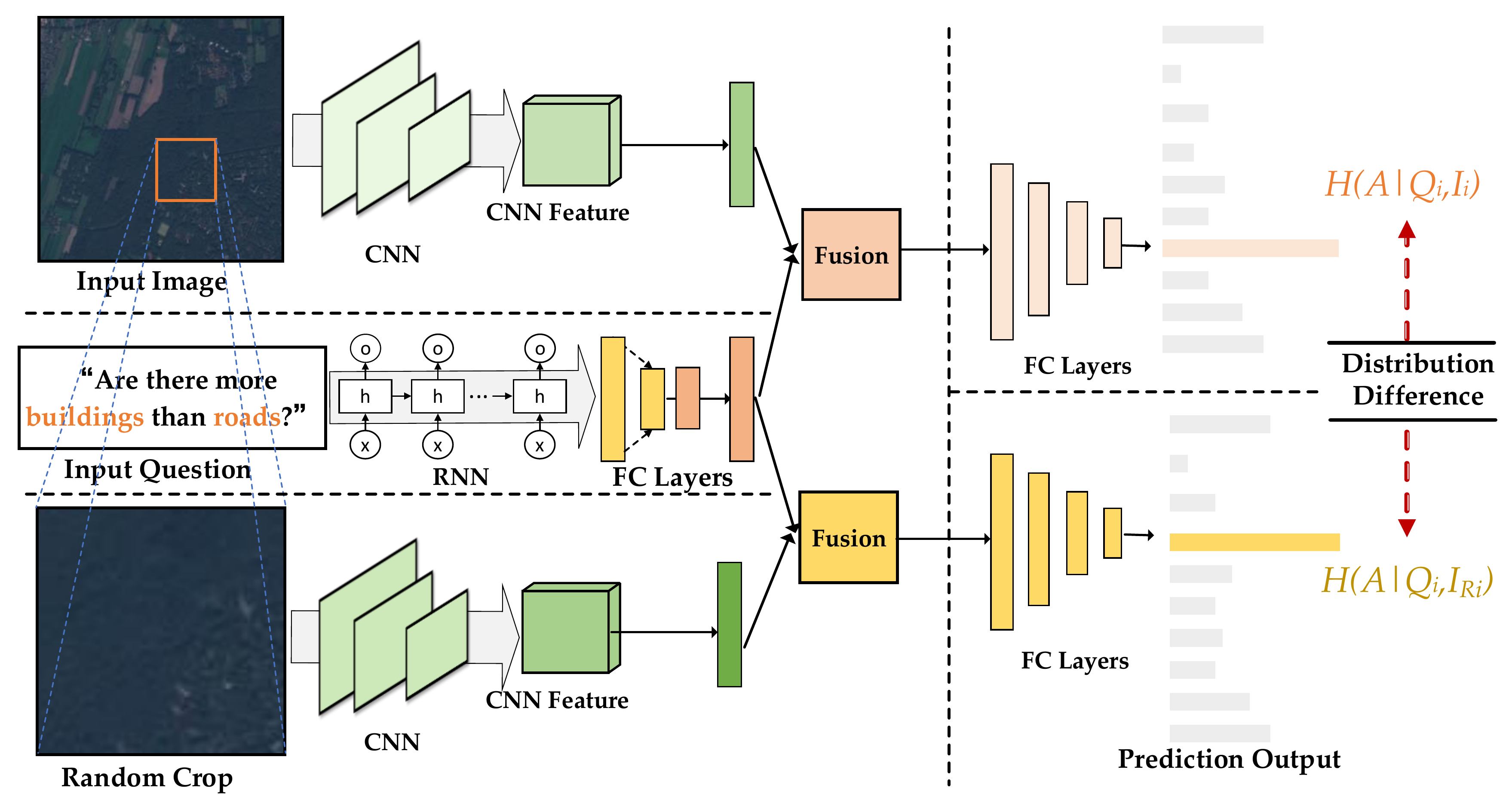}
\centering
\caption{The overview of the proposed RSVQA model. During the training stage, there are two network branches for visual feature encoding. One branch is for the original image input. The other one is an adversarial branch  with a randomly cropped patch for capturing and reducing language bias. Two regularizers are used during training for overcoming language bias. One is the adversarial branch with the negative gradient and the other is to maximize the distribution differences between the original and the adversarial branch. 
}
\label{fig1}
\end{figure*}
Despite the rapid development of RSVQA algorithms, many of them overlook the issue of language bias, which is caused by the learned superficial correlation between questions and answers \cite{kv2020reducing, yuan2021language, lao2021superficial}. For example, a VQA model may tend to answer ``yellow" when asked about the color of a banana, ignoring the actual visual content of the image. In this case, language bias will reduce the generalizability of VQA systems in real-world applications. 

Although the evaluation performance on the test set remains high, it is undesirable to blindly predict answers solely based on questions. For RSVQA tasks, the language bias problem can not be well reflected by the commonly-used evaluation metrics during the test phase. The reason is that the training and test set usually share the same question/answer distributions. While the language bias problem has been pointed out in RSVQA tasks \cite{lobry2020rsvqa}, there is still a lack of comprehensive exploration to overcome it. 
As mentioned above, language bias will make the VQA model fail to generalize to novel scenarios, significantly affecting the robustness of VQA systems.

To address this issue, in this work, we propose to reduce language bias by designing a new VQA framework with adversarial training strategies. Specifically, we introduce two regularizers to reduce language bias. One is to build an adversarial branch to prevent the model from learning language-only features for the VQA task. The other is to maximize the distribution difference between logits of the adversarial and the original branches. By introducing two regularizers during training, the proposed method can improve the robustness of the model and reduce its reliance on language features. The experimental results and analysis in this work can provide valuable insights for informing the design of future approaches aimed at reducing language bias in RSVQA tasks.

\begin{table*}[]
\centering
\caption{Performance comparisons on the LR dataset. The standard accuracies when using question and random image information are reported for both methods.}
\scalebox{0.95}{
\begin{tabular}{c|cc|cc}
\hline
\multirow{2}{*}{Question Types} & \multicolumn{2}{c|}{Easy2hard\cite{yuan2022easy}} & \multicolumn{2}{c}{Ours}                           \\ \cline{2-5} 
                                & \multicolumn{1}{c|}{Question+Image}    & Question+Random Image    & \multicolumn{1}{c|}{Question+Image} &Question+Random Image \\ \hline
Presence                        & \multicolumn{1}{c|}{0.9011}        & 0.8561               & \multicolumn{1}{c|}{0.8994}    & 0.8142            \\
Count                           & \multicolumn{1}{c|}{0.6959}        & 0.5252               & \multicolumn{1}{c|}{0.6779}    & 0.4424            \\
Comparison                            & \multicolumn{1}{c|}{0.8738}        & 0.8588               & \multicolumn{1}{c|}{0.8718}    & 0.7588            \\
Rural/Urban                     & \multicolumn{1}{c|}{0.9000}           & 0.5100                 & \multicolumn{1}{c|}{0.8600}      & 0.4300             \\ \hline
Average Accuracy                & \multicolumn{1}{c|}{0.8427}        & 0.6875               & \multicolumn{1}{c|}{0.8273}    & 0.6238            \\
Overall Accuracy                & \multicolumn{1}{c|}{0.8297}        & 0.7562               & \multicolumn{1}{c|}{0.8227}    & 0.6977            \\ \hline
\end{tabular}}
\label{tab1}
\end{table*}

\begin{table}[]
\centering
\caption{Performance comparisons with regard to language bias. The standard accuracies and performance drop when using question and random image information are reported for both methods.}
\scalebox{0.72}{
\begin{tabular}{c|ccc|ccc}
\hline
\multirow{2}{*}{Question Types} & \multicolumn{3}{c|}{Easy2hard\cite{yuan2022easy}} & \multicolumn{3}{c}{Ours}                           \\ \cline{2-7} 
                                & \multicolumn{1}{c|}{$F_a\uparrow$}    & $F_a-F_b$ &$F_m\uparrow$   & \multicolumn{1}{c|}{$F_a\uparrow$} &$F_a-F_b$ &$F_m\uparrow$ \\ \hline
Presence                        & \multicolumn{1}{c|}{0.9011}        & 0.0450       &  0.0857      & \multicolumn{1}{c|}{0.8994}    &  0.0852 & 0.1557  \\
Count                           & \multicolumn{1}{c|}{0.6959}        & 0.1707       &  0.2742      & \multicolumn{1}{c|}{0.6779}    & 0.2355  & 0.3496    \\
Comparison                            & \multicolumn{1}{c|}{0.8738}        & 0.0150      &   0.0295      & \multicolumn{1}{c|}{0.8718}    & 0.1130  & 0.2001  \\
Rural/Urban                     & \multicolumn{1}{c|}{0.9000}           & 0.3900        &   0.5442     & \multicolumn{1}{c|}{0.8600}      & 0.4300       &  0.5733     \\ \hline
Average Accuracy                & \multicolumn{1}{c|}{{0.8427}}        & 0.1552      &   0.2621      & \multicolumn{1}{c|}{0.8273}    & {0.2035}   &  {0.3267}       \\
Overall Accuracy                & \multicolumn{1}{c|}{{0.8297}}        & 0.0735      &   0.1350      & \multicolumn{1}{c|}{0.8227}   & {0.1250} & {0.2170}  \\ \hline
\end{tabular}}
\label{tab2}
\end{table}

\section{Methodology}
\label{Methodology}
In this section, we introduce the proposed framework for reducing language bias. The overview of the proposed RSVQA model is illustrated in Fig. \ref{fig1}. 
Formally, the framework takes an input image $\mathbf{x}$ and an input question $\mathbf{q}$. Initially, we utilize CNNs and RNNs to encode the image and question into deep features, respectively. Subsequently, the visual and linguistic features are fused together to obtain multi-modal representations. This computation process can be expressed by
\begin{equation}
\begin{split}
    &\mathbf{I} = f({\mathbf{x}}),\\
    &{\mathbf{Q}} = r(\mathbf{q}),\\
    &{\mathcal{F}}_1 = m(\mathbf{I},\mathbf{Q}),
\end{split}
\end{equation}
where $\mathbf{I}$ denotes visual features and $\mathbf{Q}$ denotes linguistic features. We use $f(\cdot)$ to represent CNN layers and $r(\cdot)$ to represent RNN layers. Then, multi-modal feature ${\mathcal{F}}_1$ is obtained by the feature fusion module $m(\cdot)$.

To prevent the RSVQA model from relying solely on language features, one direct approach is to amplify the influence of visual features. Building upon this concept, we incorporate an additional branch, where the original visual image is replaced with a randomly cropped patch. This random visual representation is then fused with the language features. The computation of the adversarial branch can be formally defined as
\begin{equation}
\begin{split}
    &\mathbf{I_R}= f(g(\mathbf{x})),\\
    &\mathcal{F}_2 = m({\mathbf{I_R}},\mathbf{Q}),
\end{split}
\end{equation}
where $g(\cdot)$ represents the random crop operation. Let $\mathbf{I_R}$ represent the visual features extracted from the randomly cropped image patch. Note that we employ the same CNN $f(\cdot)$ as the original branch to extract visual features. Consequently, the multi-modal feature $\mathcal{F}_2$ can be considered as the adversarial counterpart of $\mathcal{F}_1$. When comparing $\mathcal{F}_1$ and $\mathcal{F}_2$, we can observe that they share the same language input but differ in visual input. After the random crop operation, the random visual content in $\mathcal{F}_2$ can not contribute to obtaining the correct answer. As a result, the prediction of the adversarial branch should differ from that of the original branch, enabling the adversarial branch to capture the language bias. 

In order to address language bias during model optimization, we introduce a gradient reversal layer \cite{ganin2015unsupervised} following the fusion module $m(\cdot)$. By adding this layer, the original branch should update features toward a negative gradient direction to make the adversarial branch perform poorly.

Since RSVQA is modeled as a classification task, we employ multiple fully connected layers as the classifiers to predict the answer categories. As depicted in Fig. \ref{fig1}, two separate classifiers, denoted as $h_1(\cdot)$ and $h_2(\cdot)$, are utilized to predict answers for the two branches. This can be formally defined as
\begin{equation}
\begin{split}
    &\mathbf{s_1} = h_1(\mathcal{F}_1),\\
    &\mathbf{s_2} = h_2(\mathcal{F}_2),
\end{split}
\end{equation}
where $\mathbf{s_1}$ and $\mathbf{s_2}$ are the logits of the original and adversarial branches, respectively. During the training stage, the cross entropy (CE) loss is used to train the original branch, which is expressed as
\begin{equation}
\begin{split}
    &\mathcal{L}_1 = \text{CE}(\mathbf{s_1}, \mathbf{y}),
\end{split}
\end{equation}
where $\mathbf{y}$ is the ground truth label of the answer. Another regularizer is to maximize the distribution difference of logits between the original and the adversarial branches. In this way, the role of visual input can be emphasized by contrasting the multi-modal features of the randomly cropped patch and the original image. The Kullback–Leibler (KL) divergence is employed to measure the distance between the distributions of logits $s_1$ and $s_2$. Then the distribution difference loss can be computed as
\begin{equation}
\begin{split}
    &\mathcal{L}_2 = \text{KL}(\mathbf{s_1} \parallel \mathbf{s_2}),\\
    &\mathcal{L} = \mathcal{L}_1 + \lambda \mathcal{L}_2,
\end{split}
\end{equation}
where $\mathcal{L}$ is the final loss for training the model. $\lambda$ is a loss-balancing term. Note that during the inference stage, only the original branch is kept.

\section{Experiments}
\label{Experiments}
\subsection{Implementation Details}
We implement the proposed model using Pytorch. Adam optimizer with an initial learning rate of 1e-5 is utilized for training the model for 150 epochs. The batch size is set to 280. For the loss-balancing term $\lambda$, we set it to 0.1 in this work.

%Adam optimizer is 
\subsection{Evaluation Metrics}
To evaluate the performance with regard to language bias, we propose a new metric to combine the standard accuracy with the performance drop when using questions and random image information. Specifically, let $F_a$ be the accuracy of the RSVQA model evaluated on the test set when using the image-question pair as input. Suppose that $F_q$ is the accuracy of the model when using the randomly chosen image and the language information. 
On the one hand, a higher accuracy $F_a$ indicates a better model performance on the RSVQA task. On the other hand, to measure language bias, the performance drop, namely $F_a-F_q$, can be used to denote the significance of the visual input. A smaller $F_a-F_q$ means that the model relies more on the question input and the language bias of the model is large. The final metric is defined by combining these two measurements using the harmonic mean of accuracy $F_a$ and the performance drop $F_a-F_q$. The measure can be defined as 
\begin{equation}
F_m=2\times \frac{F_a^2-F_a F_q}{2F_a-F_q},
\end{equation}
where $F_m$ is the final metric for evaluating the RSVQA performance with the consideration of language bias. Note that a higher value of $F_m$ indicates a better performance.

\subsection{Results and Discussion}
We conduct experiments on the low resolution (LR) \cite{lobry2020rsvqa} dataset to validate the effectiveness of the proposed framework. As pointed out in \cite{lobry2020rsvqa}, when replacing the original image with a randomly chosen image as input, the RSVQA performance still remains high. This reveals that the LR dataset contains clear language bias. Thus, it is reasonable to choose this dataset for validating our method. For the comparison method, we take the method from \cite{yuan2022easy} as the baseline. 

The results are presented in Table \ref{tab1}. ``Question+Image'' denotes the performance when taking the original image and question as inputs. ``Question+Random Image'' stands for the performance when replacing the original image with a randomly chosen one. Table \ref{tab1} shows that clear performance drops when using randomly chosen images for both methods. When using the original input, the accuracy of \cite{yuan2022easy} is slightly higher than our method. This is primarily due to the introduced regularizers that constrain the training process, thereby enhancing generalizability, while with a slight sacrifice in RSVQA accuracy.

To compare the performance in the context of language bias, we report the results in Table \ref{tab2}. The results show that our proposed method has a larger performance drop $F_a-F_q$. Although the $F_a$ of our method is slightly lower than the baseline, the final metric $F_m$ of our method is much higher than \cite{yuan2022easy}. It indicates that our method can improve the robustness of the model by making it rely less on language features. The experimental results and analysis in this work can provide useful insights into the model design for reducing language bias in RSVQA tasks.

\section{Conclusion}
\label{Conclusion}
To improve the generalizability and robustness of the RSVQA system, we present a novel framework for reducing language bias in RSVQA tasks. The proposed framework contains an extra branch, where original images are replaced with randomly cropped image patches to enable adversarial training. Two regularizers are designed to overcome language bias during training. One is the adversarial branch with the negative gradient and the other aims to maximize the distribution differences between the original and the adversarial branches. For the performance evaluation, we design a new metric that takes the model accuracy as well as the language bias into account. Experimental results on the LR dataset show the effectiveness of our method.

\small
\bibliographystyle{ieeetr}
\bibliography{refs}

\begin{thebibliography}{10}

\bibitem{xiong2022earthnets}
Z.~Xiong, F.~Zhang, Y.~Wang, Y.~Shi, and X.~X. Zhu, ``Earthnets: Empowering ai
  in earth observation,'' {\em arXiv preprint arXiv:2210.04936}, 2022.

\bibitem{9184118}
M.~Amani, A.~Ghorbanian, S.~A. Ahmadi, M.~Kakooei, A.~Moghimi, S.~M.
  Mirmazloumi, S.~H.~A. Moghaddam, S.~Mahdavi, M.~Ghahremanloo, S.~Parsian,
  Q.~Wu, and B.~Brisco, ``Google earth engine cloud computing platform for
  remote sensing big data applications: A comprehensive review,'' {\em IEEE
  Journal of Selected Topics in Applied Earth Observations and Remote Sensing},
  vol.~13, pp.~5326--5350, 2020.

\bibitem{xiong2023gamus}
Z.~Xiong, S.~Chen, Y.~Wang, L.~Mou, and X.~X. Zhu, ``Gamus: A geometry-aware
  multi-modal semantic segmentation benchmark for remote sensing data,'' {\em
  arXiv preprint arXiv:2305.14914}, 2023.

\bibitem{yuan2023multilingual}
Z.~Yuan, L.~Mou, and X.~X. Zhu, ``Multilingual augmentation for robust visual
  question answering in remote sensing images,'' {\em arXiv preprint
  arXiv:2304.03844}, 2023.

\bibitem{antol2015vqa}
S.~Antol, A.~Agrawal, J.~Lu, M.~Mitchell, D.~Batra, C.~L. Zitnick, and
  D.~Parikh, ``Vqa: Visual question answering,'' in {\em Proceedings of the
  IEEE international conference on computer vision}, pp.~2425--2433, 2015.

\bibitem{lobry2020rsvqa}
S.~Lobry, D.~Marcos, J.~Murray, and D.~Tuia, ``{RSVQA}: Visual question
  answering for remote sensing data,'' {\em IEEE Transactions on Geoscience and
  Remote Sensing}, vol.~58, no.~12, pp.~8555--8566, 2020.

\bibitem{yuan2022easy}
Z.~Yuan, L.~Mou, Q.~Wang, and X.~X. Zhu, ``From easy to hard: Learning
  language-guided curriculum for visual question answering on remote sensing
  data,'' {\em IEEE Transactions on Geoscience and Remote Sensing}, vol.~60,
  pp.~1--11, 2022.

\bibitem{chappuis2022prompt}
C.~Chappuis, V.~Zermatten, S.~Lobry, B.~Le~Saux, and D.~Tuia, ``Prompt-{RSVQA}:
  Prompting visual context to a language model for remote sensing visual
  question answering,'' in {\em Proceedings of the IEEE/CVF Conference on
  Computer Vision and Pattern Recognition}, pp.~1372--1381, 2022.

\bibitem{yuan2022change}
Z.~Yuan, L.~Mou, Z.~Xiong, and X.~X. Zhu, ``Change detection meets visual
  question answering,'' {\em IEEE Transactions on Geoscience and Remote
  Sensing}, vol.~60, pp.~1--13, 2022.

\bibitem{kv2020reducing}
G.~Kv and A.~Mittal, ``Reducing language biases in visual question answering
  with visually-grounded question encoder,'' in {\em European Conference on
  Computer Vision}, pp.~18--34, Springer, 2020.

\bibitem{yuan2021language}
D.~Yuan, ``Language bias in visual question answering: A survey and taxonomy,''
  {\em arXiv preprint arXiv:2111.08531}, 2021.

\bibitem{lao2021superficial}
M.~Lao, Y.~Guo, Y.~Liu, W.~Chen, N.~Pu, and M.~S. Lew, ``From superficial to
  deep: Language bias driven curriculum learning for visual question
  answering,'' in {\em Proceedings of the 29th ACM International Conference on
  Multimedia}, pp.~3370--3379, 2021.

\bibitem{ganin2015unsupervised}
Y.~Ganin and V.~Lempitsky, ``Unsupervised domain adaptation by
  backpropagation,'' in {\em International conference on machine learning},
  pp.~1180--1189, PMLR, 2015.

\end{thebibliography}

\end{document}